\documentclass[10pt,twocolumn,letterpaper]{article}
\pdfoutput=1
\usepackage{iccv}
\usepackage{times}
\usepackage{graphicx}
\usepackage{amsmath}
\usepackage{amssymb}
\usepackage{booktabs}
\usepackage{comment}
\usepackage{multirow}
\usepackage{xcolor}
\usepackage{algorithm}
\usepackage{algpseudocode}
\usepackage{subfigure}
\newcommand{\algrule}[1][.2pt]{\par\vskip.5\baselineskip\hrule height #1\par\vskip.5\baselineskip}

\usepackage{comment}
\newcommand{\norm}[1]{\left\lVert#1\right\rVert}

\usepackage[pagebackref=true,breaklinks=true,colorlinks,bookmarks=false]{hyperref}
\iccvfinalcopy 
\newcommand{\parsection}[1]{\vspace{1mm}\noindent\textbf{#1}~}
\newcounter{alphasect}
\def\alphainsection{0}

\let\oldsection=\section
\def\section{%
  \ifnum\alphainsection=1%
    \addtocounter{alphasect}{1}
  \fi%
\oldsection}%

\renewcommand\thesection{%
  \ifnum\alphainsection=1%
    \Alph{alphasect}
  \else%
    \arabic{section}
  \fi%
}%

\newenvironment{alphasection}{%
  \ifnum\alphainsection=1%
    \errhelp={Let other blocks end at the beginning of the next block.}
    \errmessage{Nested Alpha section not allowed}
  \fi%
  \setcounter{alphasect}{0}
  \def\alphainsection{1}
}{%
  \setcounter{alphasect}{0}
  \def\alphainsection{0}
}%
\def\iccvPaperID{10632} 

\ificcvfinal\fi

\begin{document}

\title{Scaling Semantic Segmentation Beyond 1K Classes on a Single GPU}

\author{
Shipra Jain$^{1,2}$, Danda Paudel Pani$^{2}$, Martin Danelljan$^{2}$, Luc Van Gool$^{2,3}$\\
KTH Royal Institute of Technology, Stockholm, Sweden$^{1}$\\
Computer Vision Lab, ETH Zurich, Switzerland$^{2}$\\
KU Leuven, Belgium$^{3}$\\
{\tt\small shipra@kth.se  \{paudel,martin.danelljan,vangool\}@vision.ee.ethz.ch}
}

\maketitle
\ificcvfinal\thispagestyle{empty}\fi

\begin{abstract}
   The state-of-the-art object detection and image classification methods can perform impressively on more than 9k and 10k classes, respectively. In contrast, the number of classes in semantic segmentation datasets is relatively limited. This is not surprising when the restrictions caused by the lack of labeled data and high computation demand for segmentation are considered. In this paper, we propose a novel training methodology to train and scale the existing semantic segmentation models for a large number of semantic classes without increasing the memory overhead. In our embedding-based scalable segmentation approach, we reduce the space complexity of the segmentation model's output from O(C) to O(1), propose an approximation method for ground-truth class probability, and use it to compute cross-entropy loss. The proposed approach is general and can be adopted by any state-of-the-art segmentation model to gracefully scale it for any number of semantic classes with only one GPU. Our approach achieves similar, and in some cases, even better mIoU for Cityscapes, Pascal VOC, ADE20k, COCO-Stuff10k datasets when adopted to DeeplabV3+ model with different backbones. We demonstrate a clear benefit of our approach on a dataset with 1284 classes, bootstrapped from LVIS and COCO annotations, with almost three times better mIoU than the DeeplabV3+.
\end{abstract}
\section{Introduction}
\begin{figure}[!t]
    \centering%
    \includegraphics[width=1\linewidth]{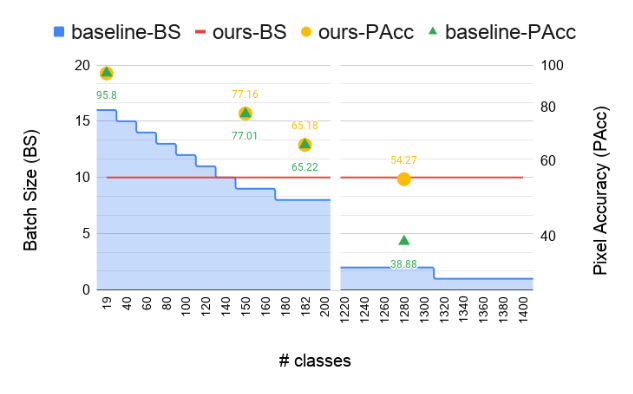}\vspace{-0.5mm} 
    \caption{The left y-axis shows the maximum batch size that can fit in a single GPU for DeepLabV3+ model vs number of classes in the dataset. The right y-axis with markers in yellow and green color shows pixel accuracy for our model and baseline for following datasets (number of classes): Cityscapes (19), ADE20k (150), COCO-Stuff10k (182) and COCO+LVIS (1284). }\vspace{-0.5mm}
    \label{fig:performance}
\end{figure} 
With the advent of deep learning, significant progress has been made in various image understanding tasks, including image classification, object detection, and image segmentation. The state-of-the-art methods can impressively classify images into 10k classes \cite{5206848} and detect 9k different objects \cite{DBLP:conf/cvpr/RedmonF17}. In contrast, segmentation models have been trained for a fairly limited number of common classes. The ability to segment a greater variety of objects, including small and rare object classes, is critical to many real-life applications like autonomous driving~\cite{beery2020synthetic} and the scene exploration~\cite{chaplot2020object}. The scaling of existing segmentation models has several unresolved challenges. One of the challenges is the unbalanced distribution of classes. As mentioned in \cite{gupta2019lvis}, due to the Zipfian distribution of classes in natural settings, there is a long tail of rare and small object classes that do not have a sufficient number of examples to train the model. The lack of segmentation datasets with a multitude of classes for learning and evaluation also limits us to develop scalable segmentation models. In fact, one can also argue from the other side. The reason for limited classes in existing segmentation datasets is the discouraging computational demand, alongside the labor-intensive annotations. The task of semantic segmentation is essentially a pixel-level classification of an image. Typically, it is performed by predicting an output tensor of $H \times W \times C$ for image size $H\times W$ and $C$ number of semantic classes \cite{DBLP:conf/cvpr/LongSD15}. This is desirable during the pixel-wise classification by employing cross-entropy loss on the  $C$-dimensional predictions. Unfortunately, the memory demand for such predictions happens to be a major bottleneck for a large number of classes, which is illustrated in Figure~\ref{fig:performance}.

Most existing works \cite{DBLP:journals/corr/abs-2005-10821,DBLP:conf/cvpr/0005DSZWTA18,9154612,DBLP:journals/corr/abs-1802-02611} primarily focus on the accuracy for datasets with a few hundred semantic classes using multiple GPUs. With the release of LVIS dataset \cite{gupta2019lvis}, efforts are being made in scaling the instance segmentation models with a large number of classes. In contrast to semantic segmentation, the task of instance segmentation is performed by classification at the region level. However, for a rich and complete understanding of the scene, semantic segmentation followed by panoptic segmentation \cite{DBLP:conf/cvpr/KirillovHGRD19} is the way to go forward. Therefore, it stands to reason that the semantic segmentation networks in the real-world will eventually have to get exposed to the classes at least high as that of classification, \ie 10K. Unfortunately, the benchmark results on ADE20k segmentation dataset with 150 classes require 4-8 GPUs during training \cite{DBLP:conf/cvpr/ZhouZPFB017}. This shows that a large number of GPUs has fueled the models for semantic segmentation. Such demand for computational resources hinders researchers in emerging economies and small-scale industries from leveraging these models for research and developing further applications.

Naive approaches for training segmentation models on large number of classes and limited GPU memory may be designed by reducing the image resolution or batch size. Such solutions regrettably
compromise the performance. As shown in \cite{DBLP:conf/cvpr/WangLZTS20}, lower resolutions (or higher strides) result in blurry boundaries and coarse predictions and miss small but essential regions, such as poles and traffic signs. 
On the other hand, \cite{DBLP:journals/ijcv/ZhouZPXFBT19} has already demonstrated the need of larger batch size to achieve state-of-the-art results. While techniques like gradient accumulation~\cite{Hermans2017AccumulatedGN} and group normalization~\cite{Wu2018GroupN} helps to reduce the effect of low batch size, but they fail to solve the problem completely when single batch size does not fit into GPU memory. When more than one GPU is available, the authors in \cite{DBLP:conf/cvpr/0005DSZWTA18} offers a promising synchronized multi-GPU batch normalization technique to increase the effective batch size. Such solutions allow scaling of classes at the cost of scaling the GPUs. However, it is interesting to look into the possibility of scaling the training for multiple classes with a single GPU, which remains unexplored.
Figure~\ref{fig:performance} also illustrates an example case: the maximum adjustable batch size of 512$\times$512 versus the number of classes, in one standard GPU (Titan XP) while training the DeepLabV3+ model with ResNet50 backbone. As expected, the batch size sharply decreases, leading to only one image per batch for 1320 classes.

In this work, we propose a novel training methodology for which the memory requirement does not increase with the number of semantic classes. To the best of our knowledge, this is the first work to study efficient training methods for semantic segmentation models beyond 1K classes. Such scaling is achieved by reducing the output channels of existing networks and learning a low dimensional embedding of semantic classes. We also propose an efficient strategy to learn and exploit such embedding for the task of semantic image segmentation. Our main motive is to improve the scalability of the existing segmentation networks, instead of competing against, by endowing them the possibility of using only one GPU during training for a very high number of semantic classes. The major contributions of this paper are summarized as follows:
\begin{itemize}
\item We propose a novel scalable approach for training semantic segmentation networks for a large number of classes using only one GPU's memory.
\item We experimentally demonstrate that the proposed method achieves 2.7x better mIoU scores on a dataset with 1284 classes, when compared against its counterpart, while retaining a competitive performance in the regime of a lower number of classes.
\item For efficiency and generalization, we introduce an approximate method to cross-entropy measure and a semantic embedding space regularization term.
\item Our method is theoretically grounded in terms of probabilistic interpretation and underlying assumptions.
\end{itemize}

\section{Related Works}
\begin{figure*}[t]
    \centering
    \includegraphics[width=1.0\linewidth ]{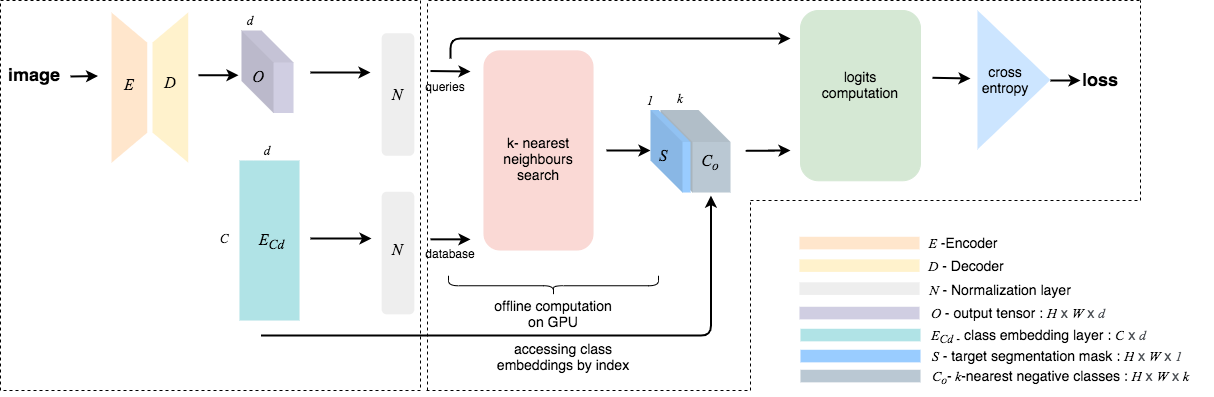} 
    \caption{\textbf{Overview}: In left, an encoder-decoder-based segmentation network $[E,D]$ with $d$-channel output (pixel embeddings) and embedding network $E_{Cd}$, followed by normalization layers $N$. In right, $k$-nearest class embeddings from $E_{Cd}$ are search for every pixel embeddings in $O$. Logits for target classes in $S$ and nearest classes in $C_o$ are computed for cross-entropy loss.}
    \label{fig:blockdiagram1}
\end{figure*}

\parsection{Efficient training for segmentation.} Existing methods are often concerned to perform segmentation in constrained devices by using limited floating point~\cite{Paszke2016ENetAD} to binary operations~\cite{zhuang2019structured} for neural networks. Other kinds are either compact by design \cite{nekrasov2019fast,DBLP:journals/corr/abs-1905-04222} or compressed after training \cite{DBLP:conf/bmvc/PoudelBLZ18,DBLP:conf/bmvc/NekrasovS018,DBLP:journals/cviu/HollidayBLKP17}. Strategies like pruning \cite{DBLP:journals/pami/LuoZZXWL19,DBLP:journals/corr/abs-2007-08386} and distilling the knowledge \cite{DBLP:journals/corr/RomeroBKCGB14,DBLP:journals/sensors/ParkH20} from the large trained model have also been explored. Almost all these approaches are either compromised in accuracy, or discount the need for high training resources \cite{carmichael2019performance}. Many works  focus on inference time on single GPU \cite{DBLP:conf/icip/Wang0LXGWL19,DBLP:conf/eccv/ZhaoQSSJ18,DBLP:journals/corr/abs-2004-02147}. Recently, \cite{DBLP:conf/cvpr/ChenJWCQ19,DBLP:journals/ijgi/ZhangLCJ19} proposed memory-efficient approaches to preserve local-global information for  high-resolution images. However, scalability issues regarding the number classes in semantic segmentation have attained little to no attention. Our method is complimentary in this regard.

\parsection{Embeddings for segmentation related tasks.} Our work is related to works that use embeddings for segmentation related tasks. Contrary to many detect and segment approaches for instance segmentation, bottom-up approaches use embeddings for one-stage training and improve performance for occluded and thin objects. A branch of work in the instance segmentation \cite{DBLP:conf/nips/NewellHD17,8014800,DBLP:journals/corr/abs-1708-02551,DBLP:journals/pami/LiangLWSYY18,DBLP:journals/corr/FathiWRWSGM17,DBLP:journals/corr/abs-1708-02550,DBLP:conf/cvpr/NevenBPG19,Kong2018RecurrentPE} trains networks for dense prediction of pixel embeddings, which are later clustered into individual instances. These methods are based on metric learning, which learns embeddings such that pixels belonging to the same instance are close to each other, and vice versa. To predict the class of instances, \cite{DBLP:conf/nips/NewellHD17,DBLP:conf/cvpr/NevenBPG19, DBLP:journals/pami/LiangLWSYY18,DBLP:journals/corr/FathiWRWSGM17, DBLP:journals/corr/abs-1708-02550} suggest to predict objectness for each object category and use cross entropy loss. \cite{8014800, DBLP:journals/corr/abs-1708-02551} compute the cluster centroids of each class over the entire training set. The classes are then inferred by comparing embeddings to the class-wise centroids. To efficiently find clustering seeds, \cite{DBLP:journals/corr/FathiWRWSGM17, DBLP:conf/cvpr/NevenBPG19} predict the heatmap for every class. To make the network end-to-end trainable, \cite{Kong2018RecurrentPE} implements a variant of mean-shift clustering using a recurrent neural network. Extensions  of these methods can be found in various applications~\cite{Athar2020STEmSegSE, DBLP:conf/cvpr/LiSVFHK18}. Differently, we exploit embeddings to capture the semantic information at the class level, unlike in the instance level of the mentioned methods. In context of semantic segmentation, \cite{chaitanya2020contrastive} used embeddings for semi-supervised segmentation, \cite{DBLP:journals/corr/HarleyDK15} refines segmentation masks using similarities between pixel embeddings and \cite{DBLP:journals/corr/HarleyDK15} learns embeddings for superpixels. \cite{Hwang2019SegSortSB} performs segmentation by extracting pixel-wise embeddings and clustering, and uses majority vote of its nearest neighbors from an annotated set to determine semantic class.

\parsection{Contrastive loss for embedding learning.} In recent years, a wide range of work \cite{DBLP:conf/iccv/DoerschGE15,DBLP:conf/cvpr/ChopraHL05,DBLP:journals/corr/abs-1906-05849,DBLP:conf/cvpr/WuXYL18} have used metric learning and contrastive losses for representation learning. Our work builds upon the same idea, which can be seen in parallel to  recently proposed contrastive cross-entropy loss in~\cite{NEURIPS2020_d89a66c7}.
In essence, \cite{NEURIPS2020_d89a66c7} is a generalization of popular triplet \cite{DBLP:journals/jmlr/WeinbergerS09} and N-pair~\cite{DBLP:conf/nips/Sohn16} losses. Contrastive losses are also very popular in self-supervised  and semi-supervised settings \cite{DBLP:journals/corr/abs-2009-00104,DBLP:journals/corr/abs-2007-13916,DBLP:journals/corr/abs-2006-08218, DBLP:journals/corr/abs-2002-05709, DBLP:conf/cvpr/He0WXG20}. Our loss fundamentally differs from the existing works, since our loss only operates on single-pixel and contrasts them against class embeddings.

\section{Embedding-based Scalable Segmentation}
 For state-of-the-art segmentation models, the output size is directly proportional to the number of semantic classes $C$. This poses a significant computational challenge while scaling them for datasets with a higher number of classes. In this work, we propose an embedding-based scalable segmentation method, which outputs a fixed number of channels and thus reduces the space complexity of output from $O(C)$ to $O(1)$. Along with the weights of the segmentation network, the model also learns $d$-dimensional class embeddings for $C$ classes. We also propose the loss functions to learn and regularize the class embeddings such that the outputs (pixel embeddings) from segmentation network for same class pixels are clustered together and are closer to their respective class embedding. An overview of the proposed method is illustrated in Figure~\ref{fig:blockdiagram1}. In the following section, we first describe the method to integrate embeddings in existing networks, then provide their probabilistic formulation followed by loss function and algorithm for loss computation.

\subsection{Low Dimensional  Embeddings}
 The key idea of our work is to reduce memory usage by representing the classes for each pixel by their corresponding embeddings. For every input image, we predict output ($O$ in Figure \ref{fig:blockdiagram1}) of size $H\times W\times d$ instead of the commonly used $H\times W\times C$, where $d<<C$. To do this, we reduce the number of filters in the last convolution layer from $C$ to $d$. In order to learn the dense target representation for every class, we add a small embedding network $E_{Cd}$ consisting of $C$  class embeddings with $d$ dimensionality. Unlike the existing models, where $C$ dimensional output at every pixel represents the pixel's classwise likelihoods, the $d$-dimensional output in our approach represents the pixel in the semantic space of class embeddings.  The embedding dimension can influence the performance of the model as with too few dimensions; the model may underfit; with too many dimensions, the model may overfit. An appropriate embedding dimension is the one to which adding further degrees of freedom would not give gains in mIoU. The reduction of dimension is followed by normalization along the depth of the output. The embedding layer is also followed by a normalization layer to ensure that embeddings lie on a unit radius hypersphere. Without normalization, a clear correlation between the length of class embeddings with the frequency of classes can be observed. 
Consistent with findings in \cite{NEURIPS2020_d89a66c7}, normalization of class and pixel embeddings helps the model suppress the bias introduced by class imbalance. 

\subsection{Probabilistic Formulation}
\label{sec:prob_formulation}
 In our approach, the distribution of pixel embeddings $O$ from the segmentation network is modeled using a gaussian mixture model. It comprises of $C$ gaussians with ${\mu_1,\mu_2,\mu_3....,\mu_C}$ centroids, identical covariance matrix $\tau I$ and equal mixing probability $\rho$, such that $C\rho$ = 1. The probability of the output embedding $x_i$ for pixel $i$ can be given by Equation \eqref{eq:px}.

\begin{equation}
    p(x_i) = \sum_{n=1}^{C} p(c_n) p(x_i|c_n)
            = \sum_{n=1}^{C}\rho \mathcal{N}(x_i|\mu_n,\tau I).
    \label{eq:px}
\end{equation}
The prior probability of class $c_n$ is $p(c_n)$. The posterior probability $p(c_n|x_n)$ gives the probability of data point $x_i$ being sampled from the gaussian of class $c_n$. As a discriminative model, segmentation network maximizes the ground truth class posterior $p(c_{y_{i}}|x_i)$. To compute the class posteriors, bayes rule is used to derive Equation \eqref{eq:pcx}.

\begin{equation}
    p(c_{y_{i}}|x_i) = \frac{p(x_i|c_{y_{i}})*p(c_{y_{i}})}{p(x_i)} = \frac{\mathcal{N}(x_i|\mu_{y_{i}},\tau I)}{\sum_{n=1}^{C} \mathcal{N}(x_i|\mu_n,\tau I)},
    \label{eq:pcx}
\end{equation}
\begin{equation}
    \mathcal{N}(x|\mu,\tau I) = \frac{1}{\sqrt{2\pi\tau}}e^{-\frac{(x - \mu)^2}{2\tau}}.
    \label{eq:normaldist}
\end{equation}
However, Equation \eqref{eq:pcx} requires computation of class-conditional probability for all classes. This makes it equally expensive in terms of computation as the $C$-channel output prediction. To overcome this problem, we propose to approximate $p(c_{y_{i}}|x_i)$ using Equation \eqref{eq:pcx_approx}.
For $x_i$, we search $k$ nearest class centroids from ${\mu_1,\mu_2,\mu_3....,\mu_C}$ denoted by $\eta(x_i,k)$ = $\{n_{1},n_{2},n_{3}....,n_{k}\}$, where $k$ $\le$ $C$. Our approach is based on the assumption that $p(c_t|x_i)$ $\approx$ 0, if $t$ $\not\in$ $\eta(x_i,k)$. The approximation error in the worst case is $\frac{1}{k} - \frac{1}{C}$, when all centroids are equidistant to $x_i$. If $k$ = $C$ or the assumption is satisfied, then the approximation error is zero.
\begin{equation}
    \overline{p}(c_{y_{i}}|x_i) 
    =\frac{\mathcal{N}(x_i|\mu_{y_{i}},\tau I)}{\sum_{n\in\eta(x_i,k)\cup y_{i}} \mathcal{N}(x_i|\mu_n,\tau I)}.
    \label{eq:pcx_approx}
\end{equation}
This probabilistic formulation motivates our loss functions described in the next section.

\subsection{Loss Functions}
\begin{figure*}[t]
    \centering
    \includegraphics[width=1.0\linewidth]{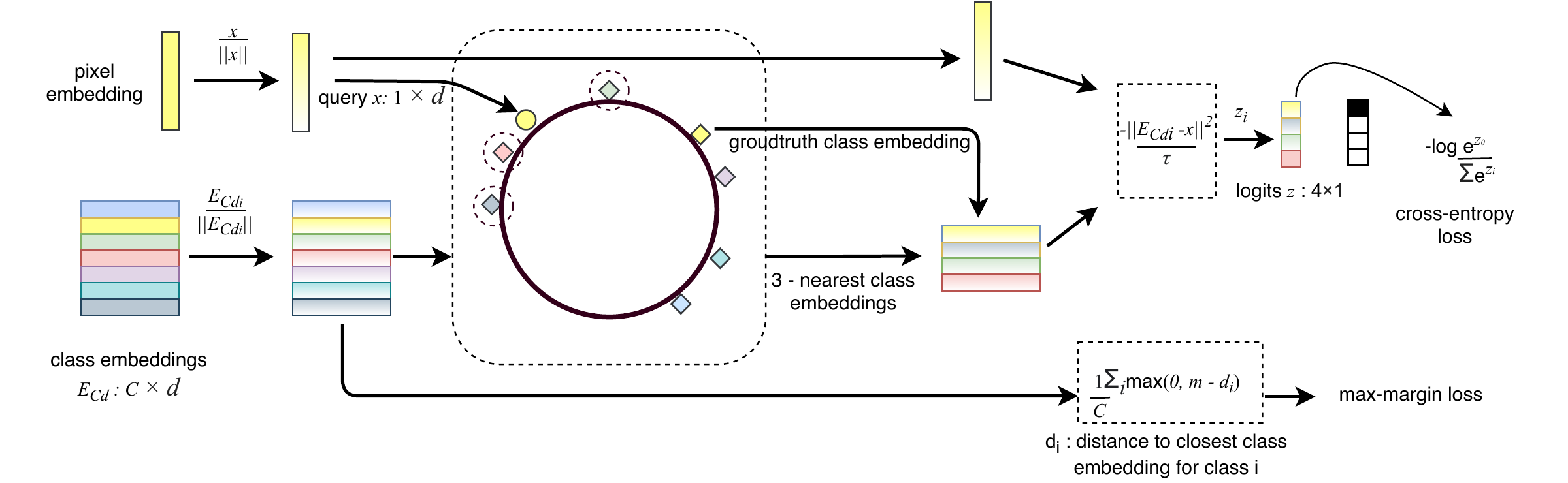} 
    \caption[Computation for Loss Function]{\textbf{Loss Computation}: A pixel embedding $x$ and class embedding $E_{Cd}$ are normalized to project on a hyperspherical manifold. For normalized $x$, $k$=3 nearest class embeddings are searched (shown by dotted circle). L2 distance between normalized $x$ and class embeddings is used to compute logits for $k$ negative nearest classes and a positive class. Further, classification and regularization loss is computed.  }
    \label{fig:blockdiagram2}
\end{figure*}
\subsubsection{Classification Loss}
The cross-entropy loss function is almost the sole choice for classification tasks in practice. It is defined as negative log-likelihood of the target class, where the class likelihood is computed from the network outputs using the softmax function. On reducing the number of channels in output, the network does not give the classwise logits directly. As shown in Equation \eqref{eq:softmax}, we use L2 distance between network outputs and class embeddings scaled by temperature $\tau$ to compute classwise logits and probability $p_i^{y_i}$ for target class $c_{y_i}$. 
\begin{equation}
p_i^{y_i} = \frac{e^{-{\|x_i - \mu_{y_i}\|^2}/{\tau}}}{\sum_{m=1}^{C}e^{-{\|x_i - \mu_m\|^2}/{\tau}}}.
\label{eq:softmax}
\end{equation}
The computation in the above equation's denominator demands a memory complexity of O($C\times D$), which does not align well with our goal. To solve this problem, we use the probabilistic formulation and assumption stated in Section \ref{sec:prob_formulation}. We propose to mine $k$ hard negative classes by searching $k$-nearest class embeddings for the pixel embedding $x_i$. In Equation \eqref{eq:approximate}, we approximate the target class probability $\overline{p}_i^{y_i}$ by using only k-nearest classes along with the target class for normalization and compute cross-entropy loss for classification. 
\begin{equation}
L = \sum_{i=1}^{N}{\log{\overline{p}_i^{y_i}}} \,,\qquad \overline{p}_i^{y_i} = \frac{e^{-{\|x_i - \mu_{y_i}\|^2}/{\tau}}}{\sum_{m\in\eta(x_i,k)\cup y_i}e^{-{\|x_i - \mu_m\|^2}/{\tau}}}.
\label{eq:approximate}
\end{equation}
The idea is to use a value of $k$ such that $O(k\times d)$ is significantly lower than $O(C)$ and can fit in the available memory. The computation of the nearest neighbours search is done in offline mode on GPU i.e.\ not included in the computational graph. The memory and speed efficient search algorithms from \cite{DBLP:journals/corr/JohnsonDJ17} can be used for this purpose.  

As cross-entropy loss maximizes target class probability, it will pull the pixel embedding closer to its target class embedding, and pixel embeddings from the same class will cluster together. Similar to previous works in \cite{DBLP:journals/corr/abs-2002-05709, NEURIPS2020_d89a66c7} , the appropriate value of temperature $\tau$ is critical for the performance of model. It represents the allowed variance across the pixel embeddings belonging to the same class and thus the compactness of clusters. 
\subsubsection{Regularization Loss}
 The classification loss models the interaction between pixel embeddings and class embeddings. To model the interaction among class embeddings and regularize them, we propose to use a max-margin loss. If class embeddings of two classes are very close, then the pixels belonging to those classes are prone to misclassification and can lead to poor generalization. The proposed loss applies repulsive force on the nearest class embedding for every class if it is closer than the margin distance $m$. Equation \eqref{eq:maxmargin} gives the regularization loss where $d_{ij}$ is the L2 distance between embeddings of class $i$ and $j$.
\begin{equation}
    L_{r} = \frac{1}{C} \sum_{i=1}^{C}\max(0,m - d_i) ±, \qquad   d_i = \min_{j \neq i, j \in C}  d_{ij} \,.
    \label{eq:maxmargin}
\end{equation}

\noindent \textbf{Learning rate scheduler.} During training, the weights for the segmentation network and the embedding network are computed and updated simultaneously. The segmentation network adjusts its weight to get pixelwise embeddings closer to corresponding class embeddings while class embeddings move closer to respective pixel embeddings. We use higher momentum and decay the learning rate of embedding network more aggressively to stabilize the training.
\subsection{The Algorithm}
We summarize the loss computation part of the proposed method in Algorithm~\ref{alg:method}. The loss computation for segmentation network $\mathsf{M_d}$ uses images $I$ with semantic masks $S$. Note that our algorithm requires an efficient GPU-compatible nearest neighbour search function represented by \textit{kNN()}, which takes a database and query vectors as inputs. Please refer to Figure~\ref{fig:blockdiagram2} for visual illustration of the algorithmic steps. The computed loss is then used to train our network illustrated in Figure~\ref{fig:blockdiagram1}.

\begin{algorithm}
 \caption{$\mathcal{L}=\textbf{LossCompute}(I,S,\mathsf{M_d},\textit{kNN()})$}
 \begin{algorithmic}[1]
 \State $O\gets \mathsf{M_d}(I)$,  \qquad $O$ shape : $B \times H \times W \times d$
 \State Turn off gradient computation
 \State   $C_{k} \gets \textit{kNN}(E_{Cd}, O$),  $C_o \gets \textit{Reshape}(C_{k}$)
 \State Turn on gradient computation
 \State $Z_o \gets \textit{Concat}(E_{Cd}(S)$, $E_{Cd}(C_o$))
 \State $Z\gets  -{\norm{O - Z_o}^2}/{\tau}$
 \State $P\gets$ \textit{Softmax($Z$)}, $P_{gt}\gets P[0]$
 \State $\mathcal{L} \gets$ \textit{mean}(-$\log{(P_{gt})}$) + ${L_{r}}$
 \State Return $\mathcal{L}$
 \end{algorithmic}
\algrule
$L_r$ is computed using Equation \eqref{eq:maxmargin}. Note that the output $O$ and class embeddings $E_{Cd}$ are normalized.
\label{alg:method}
\end{algorithm}
\vspace{-2mm}

\section{Experiments}

\noindent\textbf{Implementation Details.}
 We use DeepLabV3+ as a baseline model and the same baseline is used to investigate our approach of $d$-channel output. We use output stride as 16 and dilation rate for ASPP = [6, 12, 18]. All models are trained using the polynomial learning rate scheduler : $lr = baselr$*$(1$ - $\frac{iter}{total\_{iter}})^{power}$, the SGD optimizer with momentum, and the weight decay of 1e-4. For baseline and our segmentation network, both power and momentum are set to 0.9. These two parameters for our embedding network are set to 0.95. The base learning rate is set to 1e-2 for ADE20k, COCO-Stuff10k, and COCO+LVIS dataset and 1e-1 for Cityscapes and Pascal VOC dataset. The learning rate for the backbone is 0.1 times that of the main network and the momentum of its BN layers as 1e-2. We use margin $m$ of 0.2 in max-margin regularization loss and $\tau=0.05$. 
 The exact nearest neighbours are searched using the GPU mode of FAISS library. 

All experiments, unless mentioned, are performed using a single Titan X GPU, and the maximum possible batch size were used. For transforms, we used crop, scale, and horizontal flip in a random manner. Evaluations were performed at a single scale of the images. 
\begin{table}[b]   
\centering
\resizebox{0.9\columnwidth}{!}{%
    \begin{tabular}{cccccc}
    \toprule
    \textbf{Dataset}    &   \textbf{\# classes}    & \textbf{crop size}    &   
    \textbf{B}  &\textbf{d} &\textbf{k} \\
        \toprule
     Cityscapes  &   19    &  400 $\times$ 800  &   14/10 & 7  & 6 \\          
     Pascal VOC  &   21   &   512 $\times$ 512   &   14/10  & 7  & 8\\
     ADE20k  &   150   &   512 $\times$ 512   & 8/10  &  12  & 7  \\
     COCO-Stuff10k & 182 & 512 $\times$ 512 & 7/10&  12  & 7 \\
     COCO+LVIS & 1284 & 450 $\times$ 450 & 2/10 & 12 & 8 \\
     \bottomrule
    \end{tabular}}
    \vspace{1mm}
    \caption{\textbf{Dataset details.} Different dataset and their respective hyperparameters used to train models with ResNet50 backbone. The column \textbf{B} shows batch size for baseline and our method.}
    
\label{tab:tab9}                    
\end{table}

\parsection{Benchmark datasets.} We conducted experiments on five datasets, whose details are given in Table~\ref{tab:tab9}. The used four datasets \textbf{Cityscapes}~\cite{DBLP:conf/cvpr/CordtsORREBFRS16}, \textbf{Pascal VOC} \cite{DBLP:journals/ijcv/EveringhamGWWZ10}, \textbf{ADE20k} \cite{DBLP:conf/cvpr/ZhouZPFB017}, and \textbf{COCO-Stuff10k} \cite{8578230} are standard benchmarks. Due to lack of publicly available large scale dataset with high number of classes, we merged the COCO and LVIS dataset to demonstrate the capability of our method on 1284 classes.

\parsection{COCO+LVIS - a merged dataset.}
We build a large-scale segmentation dataset bootstrapped from stuff annotations of COCO~\cite{DBLP:conf/eccv/LinMBHPRDZ14} and instance annotation of LVIS~\cite{gupta2019lvis} for COCO 2017 images~\cite{DBLP:conf/eccv/LinMBHPRDZ14}.
LVIS is an instance segmentation dataset whose annotations are sparse for the whole image semantics. To overcome the sparsity, we merge the annotations of the stuff classes from COCO-Stuff dataset. After merging, the COCO+LVIS has the label sparsity of 19.5\% (with 18.8\% for validation). Note that, this sparsity is on par with benchmark datasets such as Pascal-MT (30.4\%)~\cite{Maninis2019AttentiveSO} and Cityscapes (28.3\%). 

We use official split of LVIS, with about 100k train and 20k validation images.
Only the semantic labels are used while ignoring the instance ids. 
LVIS has 1203 thing categories. 
Similarly, COCO has 91 stuff categories.
Between these two datasets, 10 classes are common.
This leads to the total of 1284 classes.
Labels from LVIS is prioritize over those of COCO, 
whenever they overlap. 
Please, refer to our supplementary materials for more details.
As discussed earlier, datasets with large number of classes have long-tail distributions (causing the the problem of a severe class imbalance). This is also the case for COCO+LVIS.
The mean IoU measure is known to be very sensitive to such class imbalance. Therefore, to capture a better picture, we also report frequency weighted IoU (FwIoU), along with the standard metrics: mean IoU (mIoU) and pixel accuracy (PAcc).

\subsection{Ablation Experiments}
\begin{figure}%
    \centering
    \subfigure{{\includegraphics[width=0.48\linewidth]{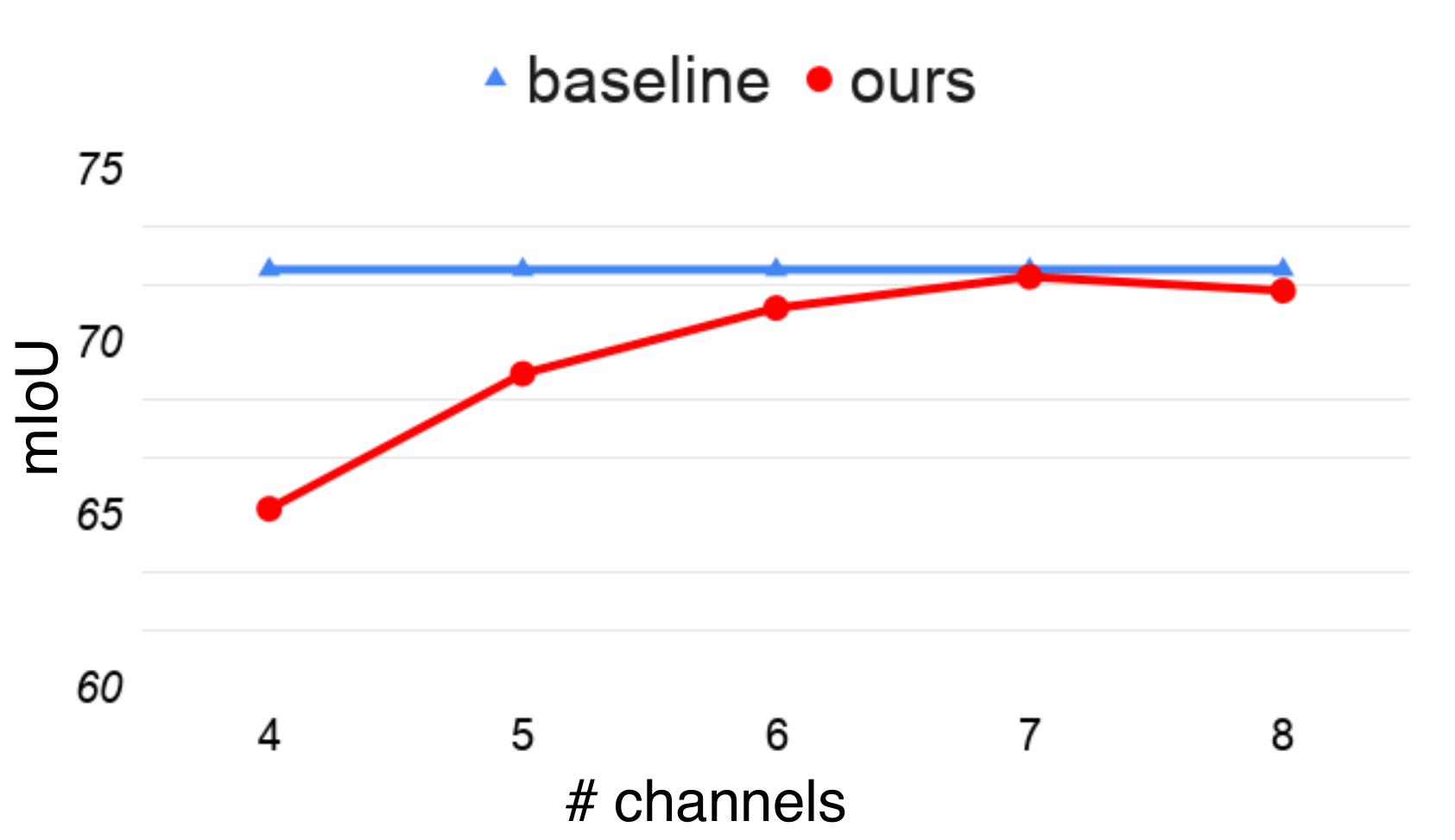} }}%
    \hspace{0.001cm}
    \subfigure{{\includegraphics[width=0.48\linewidth]{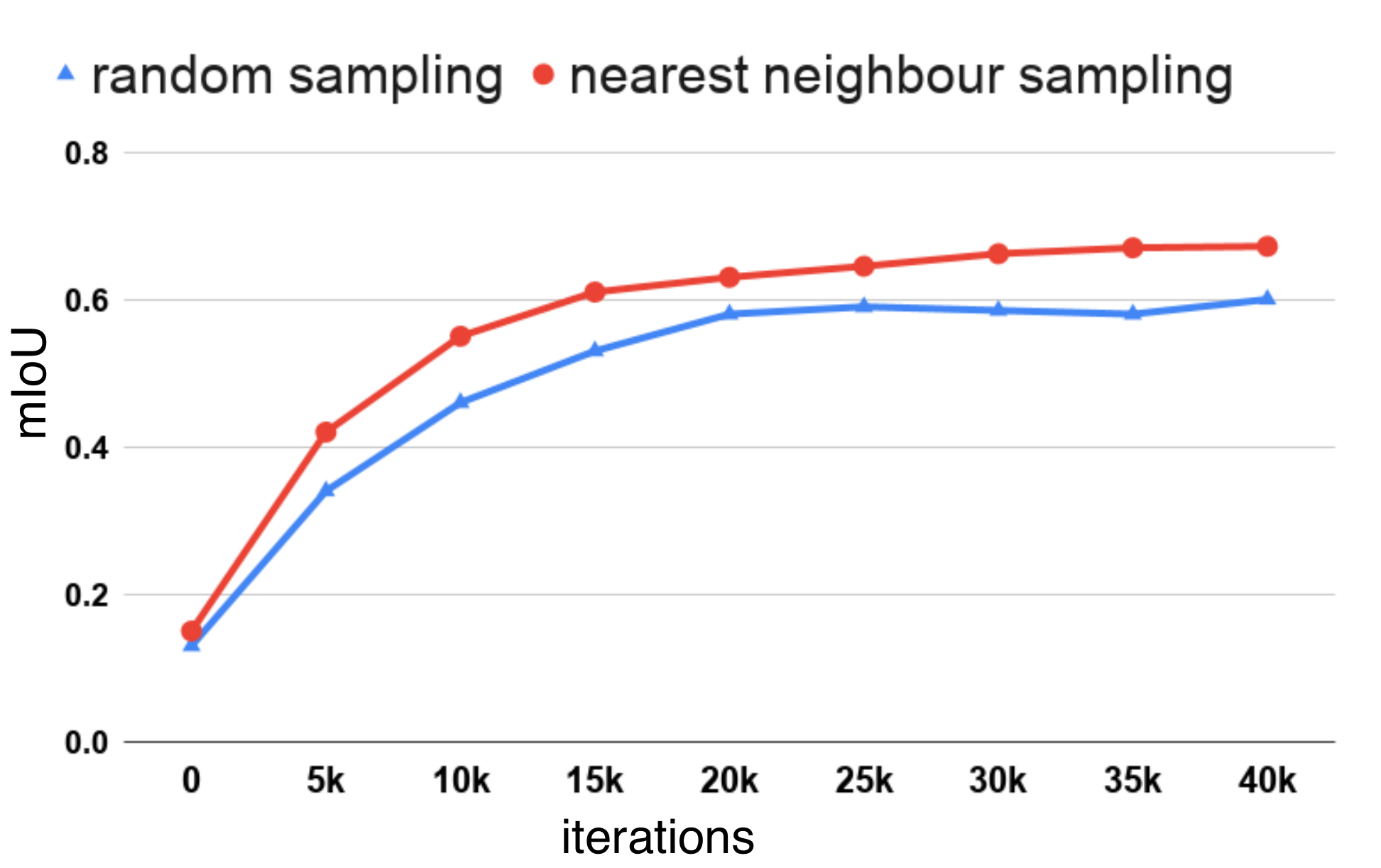} }}%
    \caption{\textbf{Number of output channels and neighbours sampling.} Number of output channels $d$ vs. mean IoU (left). The convergence of random sampling of $k$=7 vs. 7-nearest neighbours (right).}%
    \vspace{1mm}
    \label{fig:channels}%
\end{figure}

\begin{table}[b]   
\centering
\resizebox{0.7\columnwidth}{!}{%
    \begin{tabular}{cccc}
    \toprule

    \textbf{\#NN}    &   \textbf{mIoU}    & \textbf{Pixel Accuracy}    &   \textbf{iters}                \\ \midrule
     4  &   71.35    &  95.03   &   36.3k             \\          
     6  &   71.05   &   95.07   &   32.1k
  \\
     8  &   71.08   &   95.13   &   \textbf{29.1k}
           \\ \bottomrule
    \end{tabular}}\vspace{1mm}
    \caption{\textbf{Number of nearest neighbours vs. performance.} Mean IoU, pixel accuracy, and iterations for different number of nearest neighbours (\#NN). Similar performance is achieved for different \#NN with difference in convergence iterations.}\vspace{-2mm}
\label{tab:tab1}                    
\end{table}
All ablation experiments are conducted for the Cityscapes with MobileNet \cite{8578572} backbone, which are reported in Table~\ref{tab:tab1}-\ref{tab:tab2} and Figure~\ref{fig:channels}.
 
Table \ref{tab:tab1} shows that irrespective of $k$, all models converge at mIoU 71.2 $\pm$ 0.2, while the higher number of nearest neighbours being faster in convergence. Hence, $k$ can be chosen based on the trade-off between training time and the available GPU memory. These experiments do not use regularization loss. Figure \ref{fig:channels} shows the increasing performance with the increase in embedding dimension from 4 to 7, followed by a slight drop.
It also shows that the nearest neighbours offer better mIoU and convergence, compared to random sampling. 
Table \ref{tab:tab2} shows that max-margin loss provides marginal improvement in mIoU and normalization of class embeddings contributes significantly towards better mIoU.

 \begin{table}[hbt!]                           
 \centering
 
\resizebox{\columnwidth}{!}{%
    \begin{tabular}{ccccc}
    \toprule
    \textbf{nn sampling} & \textbf{normalization}    &   \textbf{max-margin loss}   & \textbf{mIoU}    &   \textbf{pixel accuracy}                   \\ \midrule
    -  & -  &    -   &  64.49   &  93.71               \\     
   $\checkmark$  & -  &   -    &  67.20   &  94.43               \\          
   $\checkmark$  & $\checkmark$  &  -  &   72.56   &   95.14
  \\ 
    $\checkmark$  & $\checkmark$  &   $\checkmark$   &  \textbf{73.03}   &  \textbf{95.40}
           \\ \bottomrule
    \end{tabular}}\vspace{1mm}
    \caption{Ablation study shows that our approach benefits from nearest neighbour (NN) sampling, normalization, and max-margin loss. Experiment in first row uses random sampling.}
    \label{tab:tab2}
\end{table}

\begin{table*}[t!]
  \begin{center}
    \resizebox{2\columnwidth}{!}{%
    \begin{tabular}{c|cc|cc|cc|cc|cc|cc|cc|ccc}
    \toprule
    \textbf{dataset} &\multicolumn{4}{c|}{\textbf{Cityscapes}} &\multicolumn{4}{c|}{\textbf{Pascal VOC}}&\multicolumn{4}{c|}{\textbf{ADE20k}}&\multicolumn{2}{c|}{\textbf{COCO-Stuff10k}}&\multicolumn{3}{c}{\textbf{COCO+LVIS}}\\
      \hline
    \textbf{backbone} & \multicolumn{2}{c|}{MobileNet}&\multicolumn{2}{c|}{ResNet50} & \multicolumn{2}{c|}{MobileNet}&\multicolumn{2}{c|}{ResNet50} & \multicolumn{2}{c|}{MobileNet} &  \multicolumn{2}{c|}{ResNet50} & \multicolumn{2}{c|}{ResNet50} & \multicolumn{3}{c}{ResNet50} \\
    \hline
    \textbf{metric}&mIoU & PAcc & mIoU & PAcc & mIoU & PAcc & mIoU & PAcc & mIoU & PAcc &  mIoU & PAcc & mIoU & PAcc & mIoU & PAcc & fwIoU\\
     \textbf{baseline} & 72.11 & 95.22 & 75.25 & \textbf{95.80} & 71.07 & 92.25 & \textbf{73.1} & \textbf{93.35} & 34.02 & 75.07 & \textbf{38.93}  & 77.01 & 32.56 & \textbf{65.22} & 1.68 & 38.88 & 22.66\\
     \textbf{ours} & \textbf{73.03} & \textbf{95.40} & \textbf{75.64} & 95.62 & \textbf{71.15} & \textbf{92.28} & 72.8 & 92.98 & \textbf{34.11} & \textbf{75.19} & 38.29 & \textbf{77.16} & \textbf{32.60} & 65.18 & \textbf{4.57} & \textbf{54.27} &\textbf{39.67} \\
     \bottomrule
    \end{tabular}}\vspace{1mm}
    \caption{Our model performs comparable to the baseline model for Cityscapes, PASCAL VOC, ADE20k and COCO-Stuff10k datasets. For COCO+LVIS dataset, it outperforms the baseline with large margin. The higher values of mean IoU (mIoU), pixel accuracy (PAcc) and Frequency weighted IoU (fwIoU) is better.}
    \label{tab:final_results} 
  \end{center}
\end{table*}

\subsection{Benchmark Results}
\paragraph{Quantitative results.}
\begin{figure}[t]
    \centering
    \includegraphics[width=1.0\linewidth]{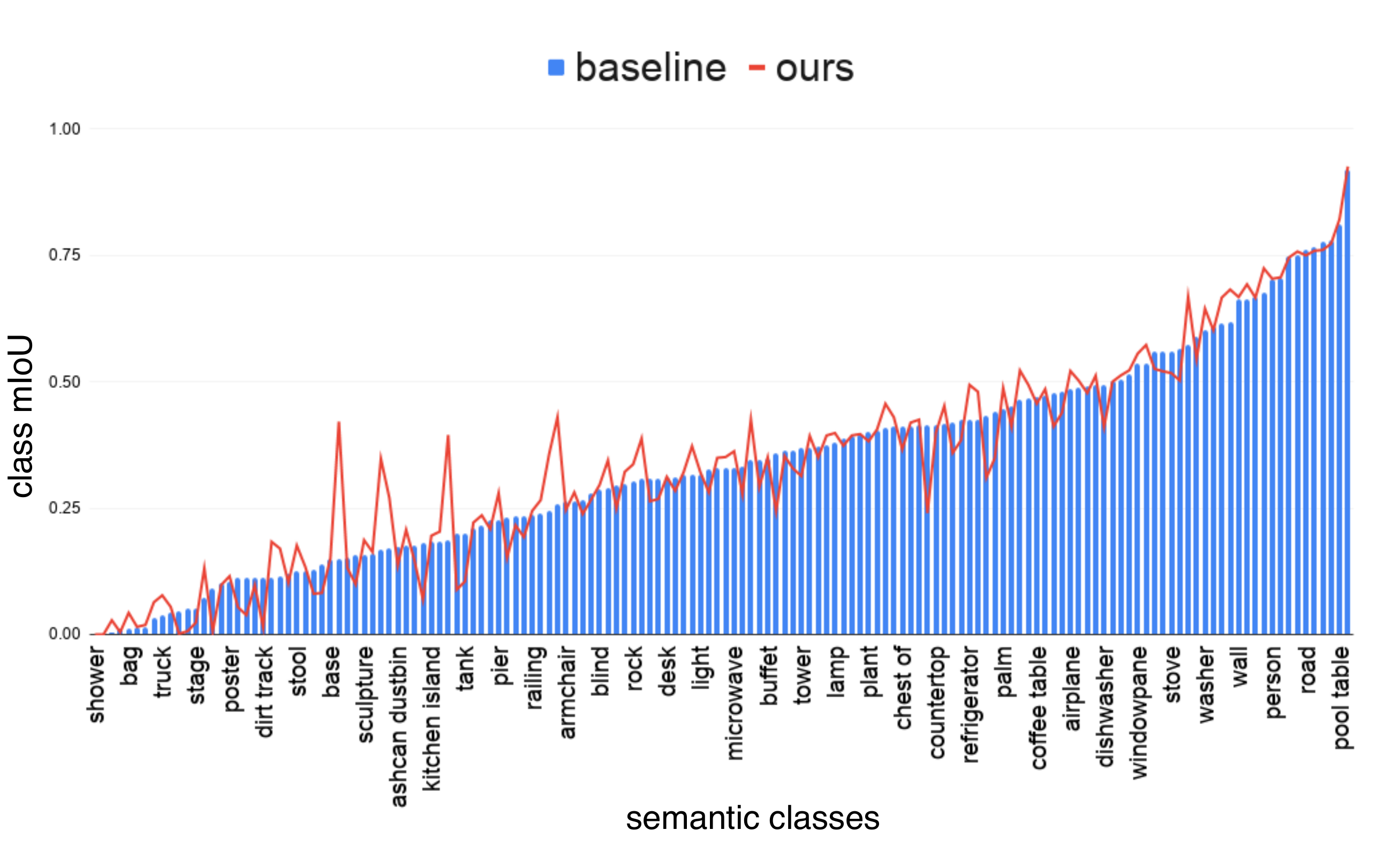} \vspace{-2mm}
    \caption[ADE20k class IoUs]{Classwise mIoU for ADE20k dataset with ResNet50 backbone. Our model does slightly better on some rare classes and performs comparable on dominant classes.}
    \label{fig:classious_ade20k}
    \vspace{-2mm}
\end{figure}
For datasets with a lower number of classes, Table \ref{tab:final_results} shows that the performance of our model with both ResNet50 and MobileNet backbones is comparable to that of the baseline.  In ADE20k, the distribution of classes is highly unbalanced as the stuff classes like ‘wall’, ‘building’, ‘floor’, and ‘sky’ occupy more than 40\% of the annotated pixels. In contrast, the discrete objects, such as ‘vase’ and ‘microwave’ at the tail of the distribution, occupy only 0.03\%  of the annotated pixels. Figure~\ref{fig:classious_ade20k} shows a comparison between classwise IoUs of ADE20k for both models. Classes in the plot are sorted based on IoUs for the baseline. We observe that our model performs better for some rare classes like shower, apparel, and stool over baseline, and these classes occur in some specific context like bathroom or bedroom. We hypothesize that our learned embeddings allows rare classes to implicitly borrow knowledge from the associated semantic context. As the frequency of classes increases, both models perform similarly. 

\begin{figure}[b]
    \centering
    \includegraphics[width=0.9\linewidth]{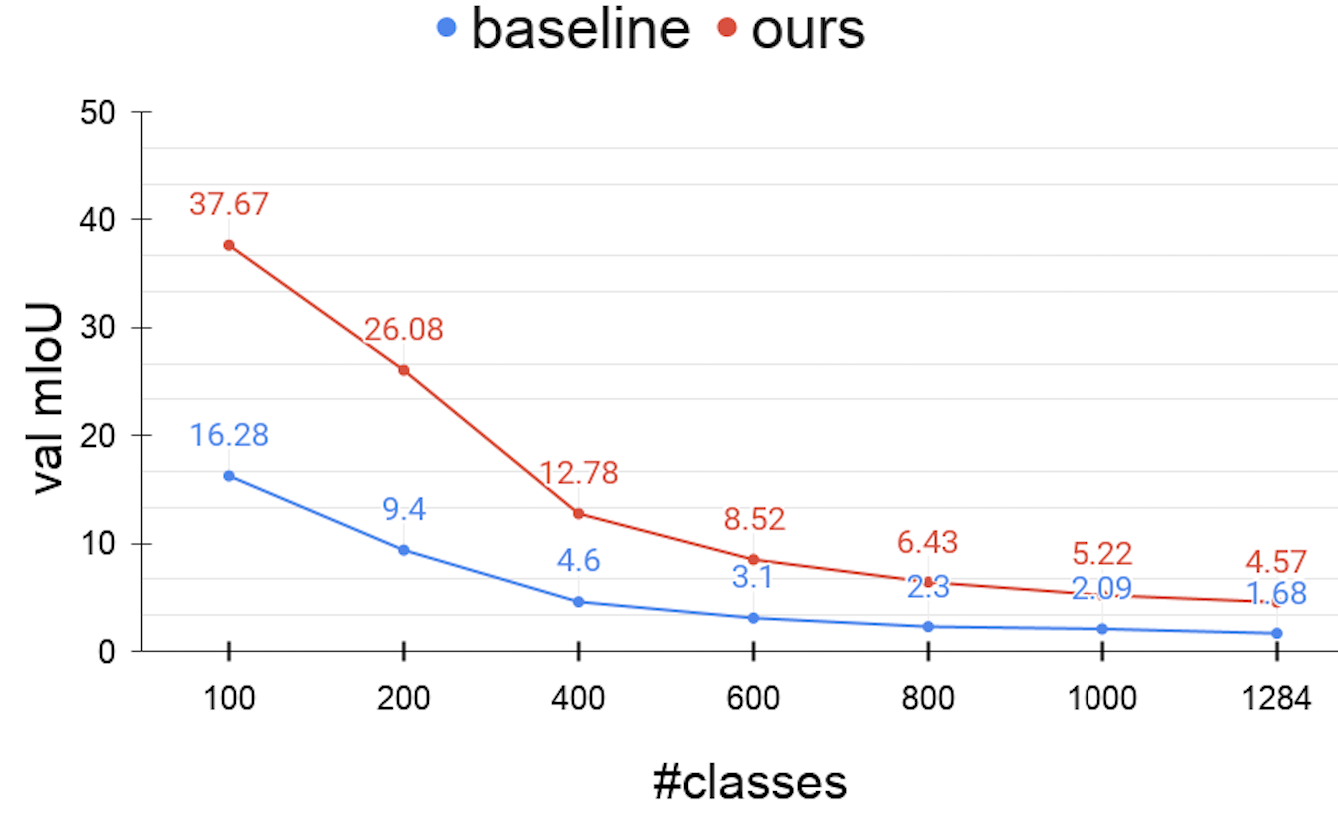} \vspace{-1mm}
    \caption[COCO+LVIS]{mIoU on COCO+LVIS with increasing number of classes, with most frequent first, for baseline and our method.}
    \label{fig:classious_lvis}
    \vspace{-2mm}
\end{figure}

For the COCO+LVIS dataset, our model clearly outperforms the baseline in terms of both mIoU and pixel accuracy. The low mIoU for both models, when compared to other datasets, can be explained by the long tail of thing classes in LVIS annotations. Figure \ref{fig:classious_lvis} shows that as we increase number of rare classes, mIoU drops. Among 1284 classes, 220 classes occur in less than ten images in the training dataset. Please recall, the challenge of class imbalance is out of the scope of our work.
To provide a complete picture, we also report  the frequency weighted IoU for COCO+LVIS. The superior performance of our method for the COCO+LVIS can be explained by the five times better batch size that it can fit in a single GPU. Lower batch size leads to noisy estimation of batch statistics in BatchNorm layer. To reduce the effect of low batch size in baseline model, we perform experiments using gradient accumulation (GA)~\cite{Hermans2017AccumulatedGN} and group normalization (GN)~\cite{Wu2018GroupN}. Table \ref{tab:tab10} shows that GA and GN help to improve the performance of both the models. GA increases the effective batch size of all the layers in network except BatchNorm as the mean and variance for every batch are computed during the forward pass. GN makes the computation of mean and variance independent of batch size. However, these techniques are not the substitute for our approach as our major contribution lies on restricting the number of output channels, thus decreasing the memory complexity from $O(C)$ to $O(1)$. Using GN/GA (with baseline model) alone would not be possible for very high number classes or larger images as even a single image would not fit into the memory (because of $O(C)$ complexity).
\begin{table}[t]
  \begin{center}
    \resizebox{0.6\columnwidth}{!}{%
    \begin{tabular}{c|c|c|c}
    \toprule
      \textbf{model} & \textbf{mIoU} & \textbf{FwIoU} & \textbf{PAcc}\\
      \hline
      baseline & 1.68 & 22.66 & 38.88 \\
      ours & \textbf{4.57} & \textbf{39.67} & \textbf{54.27} \\
     \hline
     baseline + GA & 2.76 & 29.57 & 46.34 \\
      our + GA & \textbf{5.01} & \textbf{41.87} & \textbf{57.05} \\
     \hline
     baseline + GN & 5.15  & 37.89 & 53.45 \\
      ours + GN & \textbf{6.26}  & \textbf{43.03} & \textbf{59.01}
      \\
      
     \bottomrule
    \end{tabular}}\vspace{1mm}
    \caption{Results on COCO+LVIS dataset with Gradient Accumulation (GA) and GroupNorm (GN). Gradients are accumulated over 5 and 2 steps for baseline and ours, respectively. In GN experiments, we have used groups of 16-channels for both methods.}\vspace{-2mm}
      \label{tab:tab10} 
  \end{center}
  \vspace{-2mm}
\end{table}
\begin{table}[t]
  \begin{center}
    \resizebox{0.8\columnwidth}{!}{%
    \begin{tabular}{cccc}
    \toprule
      \textbf{dataset} & \textbf{model} & \textbf{train BS} & \textbf{memory (in GB)}\\
      \hline
      \multirow{2}{*}{Cityscapes}& baseline & \textbf{14} & 12.1 \\
      \cline{2-4}
      &ours &12 &10.4 \\
     \hline
     \multirow{2}{*}{ADE20k}& baseline & 8 & 10.3 \\
     \cline{2-4}
      &ours &\textbf{10} & 10.0 \\
     \hline
     \multirow{2}{*}{COCO+LVIS}& baseline & 2 & 9.94 \\
     \cline{2-4}
      &ours &\textbf{10} & 10.4
      \\
     \bottomrule
    \end{tabular}}\vspace{1mm}
    \caption{Analysis of peak GPU memory usage and maximum batchsize for 1 GPU. For Cityscapes dataset, baseline has better memory consumption while our model is memory efficient for ADE20k and COCO+LVIS datasets.}\vspace{-2mm}
      \label{tab:tab5} 
  \end{center}
  \vspace{-2mm}
\end{table}

\begin{figure}[t]
    \centering
    \includegraphics[width=1.0\linewidth]{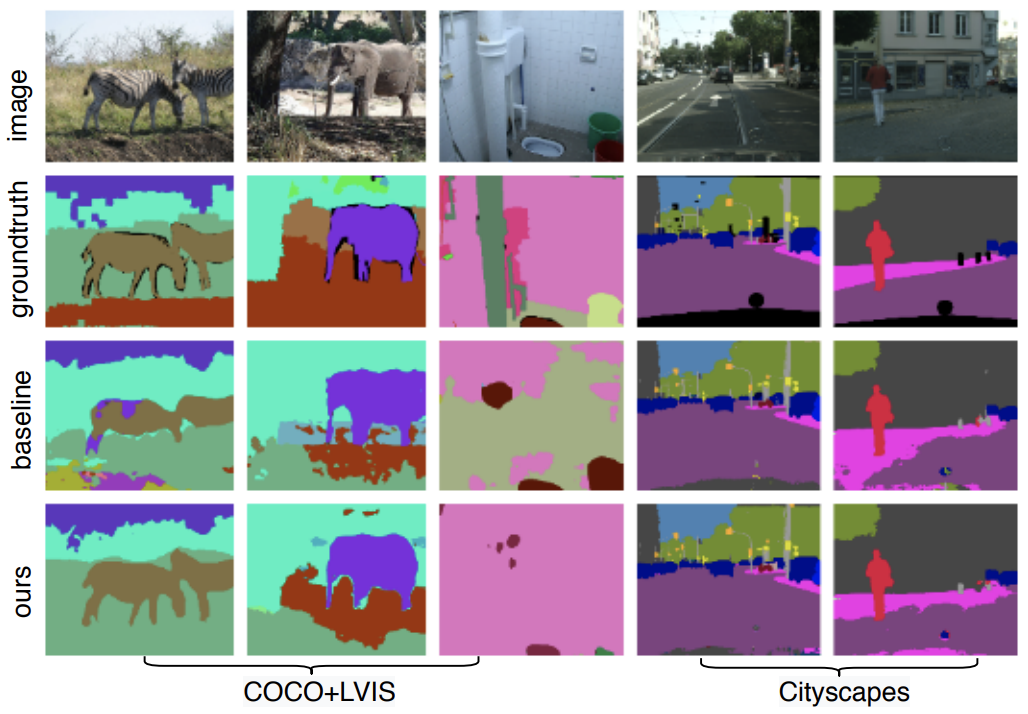}
    \vspace{-2mm}
    \caption[]{Qualitative results of our method and the baseline. Black color denotes the unlabelled pixels. For COCO+LVIS dataset, both models miss rare classes such as bucket and pipe. Our model performs better than baseline for dominant classes like wall. For Cityscapes, both model provide similar results.}
    \label{fig:QR1}
    \vspace{-2mm}
\end{figure}

\parsection{Analysis of memory consumption.}
In Table \ref{tab:tab5}, we investigate the peak memory usage in GPU during training. We observe that for smaller datasets like Cityscapes, baseline uses less memory to accommodate the bigger batch size. However, our approach is better suited for datasets with a higher number of classes like ADE20k.  In this case, our approach accommodates a bigger batch size for the same memory. Despite of any increase in number of classes, our model's memory requirement remains almost the same, thanks to the $O(1)$ complexity of the proposed method. This allow us to scale to 1k+ classes and still use the batch size of 10. On the other hand, the baseline model can only fit a batch size of 2 in a single GPU. For details on inference time, please refer to supplementary material.
\begin{figure}[t]
    \centering
    \includegraphics[width=1.0\linewidth]{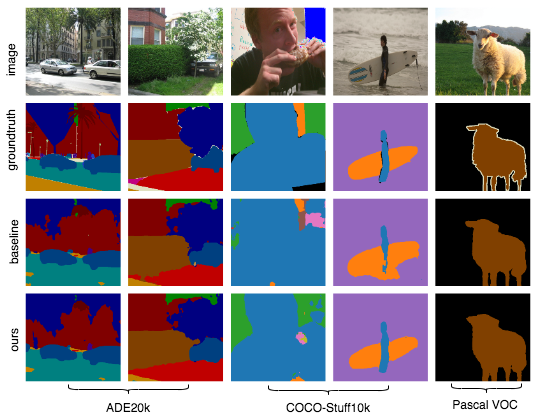}
    \vspace{-2mm}
    \caption[]{Qualitative results for our method and the baseline. For ADE20k, COCO-Stuff10k, and Pascal VOC datasets both the models provide similar qualitative results.}
    \label{fig:QR2}
    \vspace{-2mm}
\end{figure}

\parsection{Qualitative results.}
In Figure \ref{fig:QR1} and \ref{fig:QR2}, we show qualitative results. In COCO+LVIS dataset, rare and small area classes are mostly missed by both the models. This also reflects in lower mIoU scores. Our model segments the dominant classes like wall and grass much better than the baseline. For CityScapes, Pascal VOC, ADE20k, and COCO-Stuff10k, segmentation masks from both models look very similar. We also notice that almost same set of pixels are misclassified by both the models in many examples.

\begin{figure}[t]
    \centering
    \includegraphics[width=1.0\linewidth]{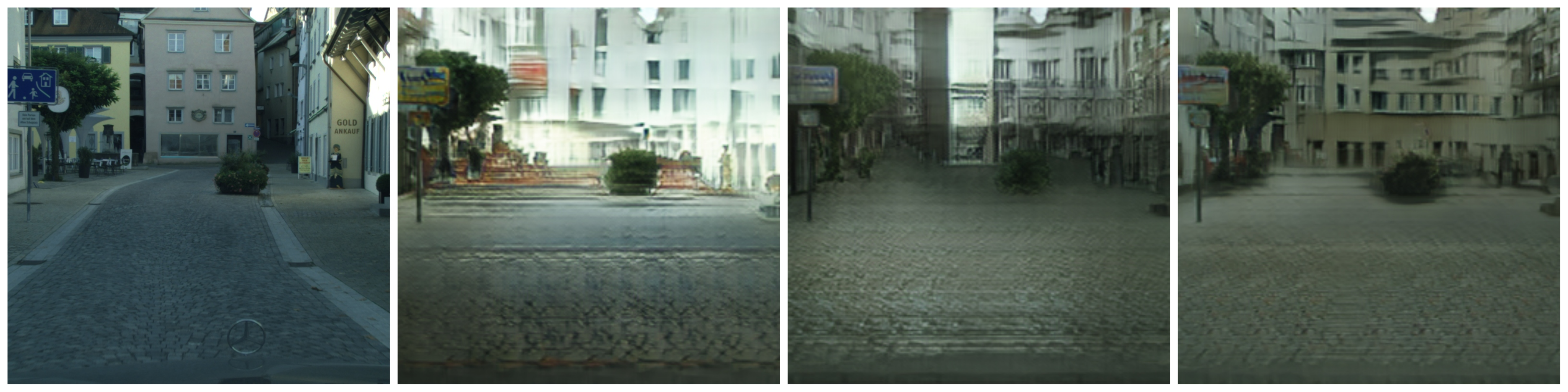}
    \vspace{-2mm}
    \caption[]{Synthesized images for Cityscapes. Left to right: real image;  generated using: one-hot encoding (FID = \textbf{60.47}); random embeddings (FID = \textbf{64.14}); our class embeddings (FID = \textbf{58.34}).}
    \label{fig:spade}
    \vspace{-2mm}
\end{figure}

\parsection{Semantic class embeddings for image synthesis.}
Using the learned class embeddings, our method performs well for the task of semantic segmentation. This suggests that our embeddings capture the semantics of the classes and represent them efficiently in lower-dimensional space. In order to demonstrate the utility, beyond segmentation, of our learned embeddings, we conducted experiments with SPADE network \cite{DBLP:conf/cvpr/Park0WZ19} to synthesize photo-realistic images. SPADE takes class semantics in the form of a one-hot vector corresponding to the class label for every pixel as input. We conduct three experiments : 1. one-hot vector semantics (19 classes) as input with $B=3$, 2. randomly initialized 7-dim embeddings as input with $B=4$, and 3. 7-dim class embeddings from our trained segmentation network with $B=4$.
 Figure \ref{fig:spade} shows image examples generated for the Cityscapes test dataset using a single GPU. Our embeddings achieve a lower FID score than random embeddings, which suggests that our learned class embeddings can also be used for synthesis. Embedding-based semantic inputs for the memory-efficient generation of images, with a higher number of classes, remains a promising direction for future work. For visualization of our class embeddings, please refer to the supplementary materials.

\section{Conclusions}
In this work, we address the problem of increase in the memory complexity of existing segmentation approaches with increasing number of semantic classes. By leveraging our understanding of metric learning and probabilistic mixture models, we proposed a novel approach to train the segmentation models. The proposed method can be used for any number of classes to train the semantic segmentation networks in a single GPU's memory. Our experiments demonstrate that the proposed method can retain the performance, while improving the scalability; thus allowing us to segment a large number of classes. Our work raises the memory issues with the existing methods, and proposes a new research direction to perform large scale segmentation in a meaningful way.

{\small
\bibliographystyle{ieee_fullname}
\bibliography{main}
}
\newpage
\begin{alphasection}
\begin{figure*}[htb!]
    \centering
    \includegraphics[width=1.0\linewidth]{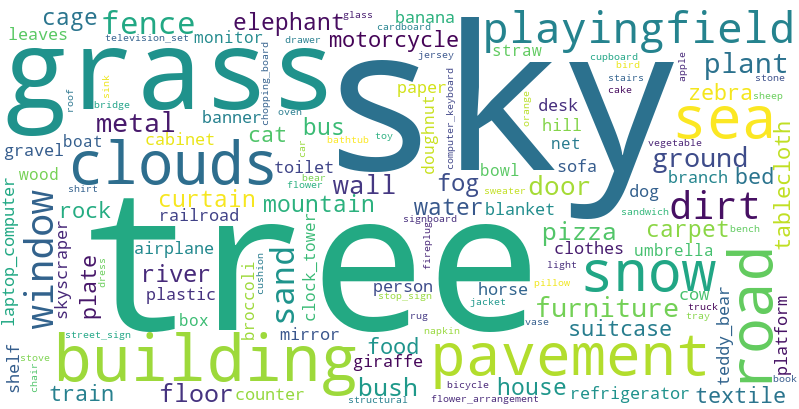}
    \vspace{-2mm}
    \caption[wordcloud]{Word cloud of semantic classes of COCO+LVIS dataset. Bigger font size means higher pixel ratio.}
    \label{fig:wordcloud}
    \vspace{-2mm}
\end{figure*}

\begin{figure}[htb!]
    \centering
    \includegraphics[width=1.0\linewidth]{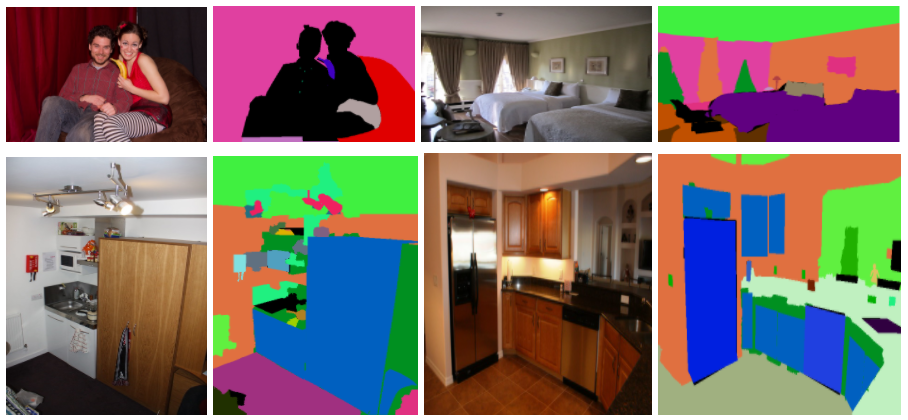}
    \vspace{-2mm}
    \caption[common classes]{Top: Curtain as stuff class (in left) and as thing class (in right). Bottom: Cabinet as stuff class (in left) and as thing class (in right).}
    \label{fig:commonclasses}
    \vspace{-2mm}
\end{figure}

\begin{figure}[htb!]
    \centering
    \includegraphics[width=1.0\linewidth]{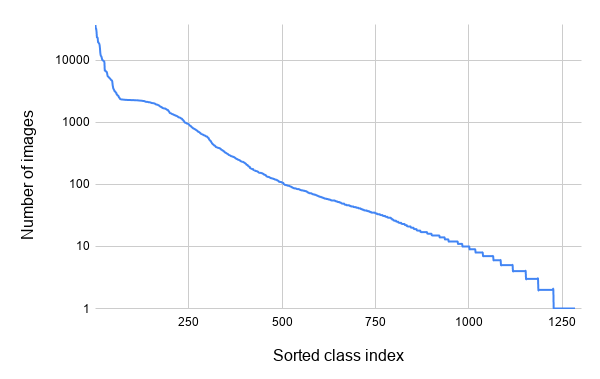}
    \vspace{-2mm}
    \caption[numimages]{Number of images per semantic class.}
    \label{fig:numimages}
    \vspace{-2mm}
\end{figure}
\begin{figure*}[t]
    \centering
    \includegraphics[width=1.0\linewidth]{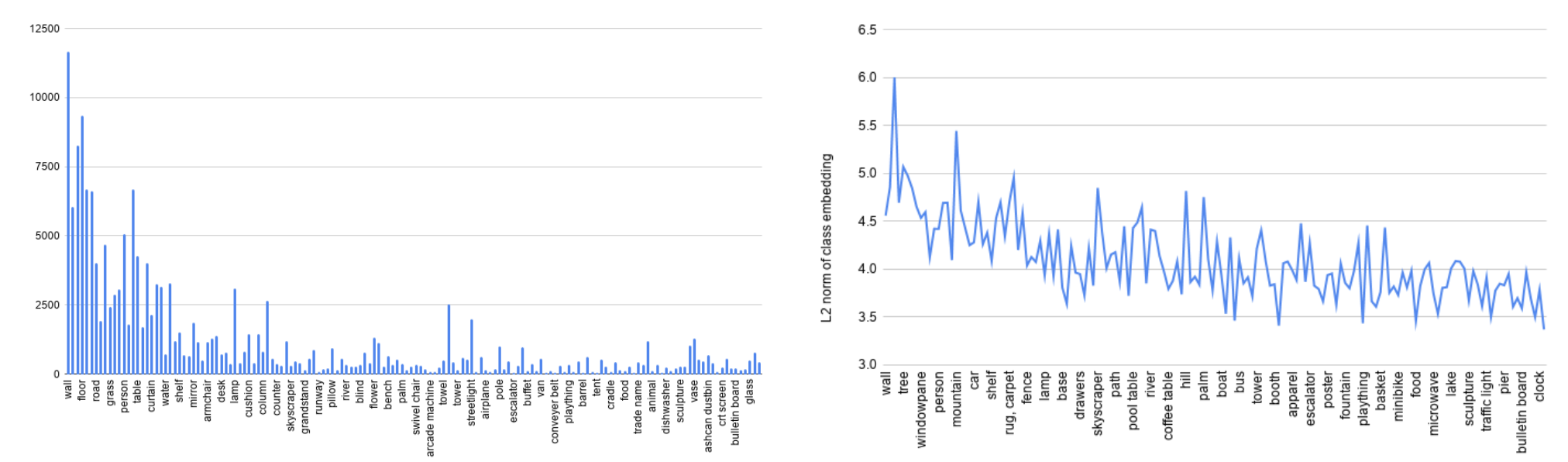}
    \vspace{-2mm}
    \caption[]{\textbf{Class Frequency and Embedding length} Left: Frequency of classes in training dataset for ADE20k dataset. Right: Length of class embeddings when trained the model without normalization layers. There is a correlation between frequency of class and distance of its class embedding from origin.}
    \label{fig:norm}
    \vspace{-2mm}
\end{figure*}
\begin{figure}[htb!]
    \centering
    \includegraphics[width=1.0\linewidth]{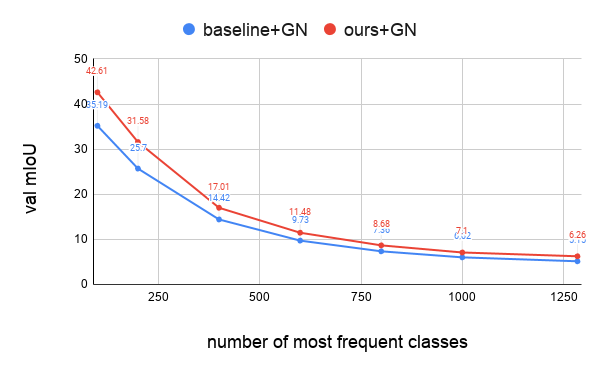}
    \vspace{-2mm}
    \caption[GN]{mIoU for COCO+LVIS dataset with increasing number of classes, with most frequent first.}
    \label{fig:GNmious}
    \vspace{-2mm}
\end{figure}

In the supplementary material, we first report more details on COCO+LVIS dataset and its experiments, discuss the training and inference time of our approach, followed by visualization of semantic class embeddings and more qualitative results. We have also attached a zip file which has our source code and script to build COCO+LVIS dataset.
\section{More details about COCO+LVIS}
The vocabulary of our bootstrapped COCO+LVIS dataset is build from 181 stuff and 1203 thing classes from COCO and LVIS annotations, respectively. As stuff classes can sometimes be things and vice-versa depending upon the scene and context, there is an overlap of 10 classes between COCO and LVIS vocabulary. The common classes are pillow, curtain, table, cabinet, banner, towel, salad, napkin, blanket, and cupboard. Figure~\ref{fig:commonclasses} shows examples of classes considered as stuff and thing depending upon the context. COCO+LVIS has 1284 classes. Figure~\ref{fig:wordcloud} shows the class names and their font size determined by pixel ratio. Figure~\ref{fig:numimages} shows the number of images every class occur in and the long-tail distribution of classes. 

We also evaluate our and baseline model with GN for n-most frequent classes in Figure~\ref{fig:GNmious}. It shows that our model clearly outperforms the baseline, even when Group Normalization is used. To understand the performed loss for LVIS+COCO incurred by limited computational resources, we also run our model (no GN) on 4 GPUs (16 GB each) with synchronized batch norm. With batch size 40 (10 per GPU) and embedding size 16, we achieve \textbf{mIoU 8.78}. We did not perform hyperparameter search for this experiment. We believe that by reducing the batch size to reasonable number such as 16 or 24 while increasing embedding dimension would help us further improve mIoU.
\section{Training and Inference time}
\begin{table}[b]
  \begin{center}
    \resizebox{0.8\columnwidth}{!}{%
    \begin{tabular}{cccc}
    \toprule
      \textbf{dataset} & \textbf{model} & \textbf{inference time} & \textbf{training time}\\
      \hline
      \multirow{2}{*}{Cityscapes}& baseline & 0.195 & 4.94 \\
      \cline{2-4}
      &ours &0.233 & 5.34 \\
     \hline
     \multirow{2}{*}{ADE20k}& baseline & 0.023 & 3.01 \\
     \cline{2-4}
      &ours &0.026 & 4.80 \\
     \hline
     \multirow{2}{*}{COCO+LVIS}& baseline & 0.049 & 4.20 \\
     \cline{2-4}
      &ours &0.036 & 5.06
      \\
     \bottomrule
    \end{tabular}}
    \caption{Analysis of inference and training time. Inference time is given in seconds per image and training time is given in seconds per iteration. Lower training and inference time is better. }\vspace{-2mm}
      \label{tab:tab1} 
  \end{center}
  \vspace{-2mm}
\end{table}
 Table \ref{tab:tab1} shows inference and training times for datasets with a different number of semantic classes. In comparison to the baseline model, our model takes slightly higher inference time for datasets with a lower number of semantic categories and lower inference time for datasets with higher number of classes. During inference, we use index functionality of FAISS library, which first builds an index using class embeddings and then another function call is used to perform the nearest neighbour search. The inference time in our computation includes the duration of the forward pass and segmentation prediction and does not include time for model initialization. We compute inference time for models with ResNet50 backbone and use maximum validation batch size that can fit in GPU. Images with 1024 $\times$ 2048 resolution for Cityscapes dataset and 512 $\times$ 512 for ADE20k and COCO+LVIS dataset are used.

We train models for 200, 80 and 40 epochs for Cityscapes, ADE20k and COCO+LVIS datasets respectively. In terms of training time, our model takes higher time per iteration for all datasets. It is one of the significant weakness of our approach. $k$-nearest search computation performed for every pixel is the major bottleneck in our computation time. This computation can be optimized using non-exhaustive search methods like clustering the class embeddings and searching the neighbours for the query only in the cluster in which it lies. We notice that output pixel embeddings for adjacent pixels are very close in embedding space, and this property can be used to compute nearest neighbour search for only 0.25 or 0.125 fractions of total pixels and use same negative samples for 4 or 8 neighbouring pixels. While training both the models for the same number of epochs, we also notice that the baseline converges in 2-6 fewer epochs than our model. This depends upon the number of nearest neighbours $k$ used during the training.


\section{Semantic Embeddings and Visualizations}
In this section, we investigate the relation between embeddings of different pixels in an image and the class embeddings learned by our model. Figure \ref{fig:norm} shows the correlation between the frequency of classes and length of class embedding when normalization layers are not used. Therefore, normalization is essential to suppress the bias caused due to the class imbalance. Figure \ref{fig:pix_emb} shows an example of ground truth segmentation mask from ADE20k dataset and corresponding pixel embeddings from our model projected in 2D space. As desired, the pixels belonging to the same class (with the same color) are clustered together. We also notice that the transition of embedding from one class pixel to an adjacent pixel of another class is smooth. This nature of our pixel embeddings might lead to misclassification of pixels at the boundary of the object. Figure \ref{fig:RGB_embed} shows examples of predicted masks and projection of their pixel embeddings to RGB space. The same colour of pixels in the projection image suggests that their pixel embeddings are closer in feature space, but their nearest class embedding can be different (can be seen from predicted masks). 
\begin{figure}[htb!]
    \centering
    \includegraphics[width=1.0\linewidth]{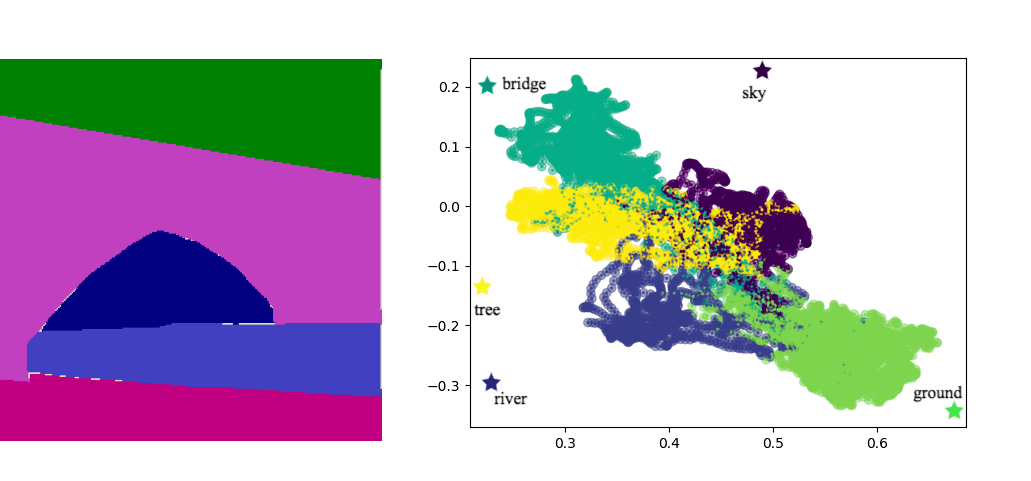}
    \vspace{-2mm}
    \caption[]{\textbf{Pixel Embeddings in 2D space} Left: Example of groundtruth segmentation mask from ADE20k dataset. Right: Circle-shaped markers - pixel embeddings, output from our model is projected into 2D. Star-shaped markers- class embeddings. The color of circular marker denotes the target class of pixel. }
    \label{fig:pix_emb}
    \vspace{-2mm}
\end{figure}

We perform agglomerative clustering of classes based on the class embeddings learned by our model in Figure \ref{fig:ade20k_clusters} and \ref{fig:lvis}. We notice in Figure~\ref{fig:ade20k_clusters} that classes which occur in a similar context or are semantically similar are closer in feature space. There are several small sub-trees for different contexts like kitchen, scenery, bedroom, interior and many more. For example, pillow, cushion, bed, couch, stool, chair and hassock are clustered together. Also, kitchen equipment like microwave, refrigerator, cabinet, dishwashing machine, cooking stove, sink, kitchen island and countertop fall in same sub-tree. Semantically similar classes like monitoring device and CRT screen are adjacent. The light source and lamp is another pair of adjacent classes with the same semantics. In Figure \ref{fig:lvis}, we perform agglomerative clustering for the hundred most frequent classes from COCO+LVIS dataset. We observe similar clusters for COCO+LVIS dataset also. We also performed k-means clustering on embeddings from ours+GN model and we have attached the list of 70 clusters in supplementary. Clusters such as (bear, grizzly, polar\_bear) and (cup, mug, teacup) suggests that embeddings are semantically meaningful.

\begin{figure}[htb!]
    \centering
    \includegraphics[width=1.0\linewidth]{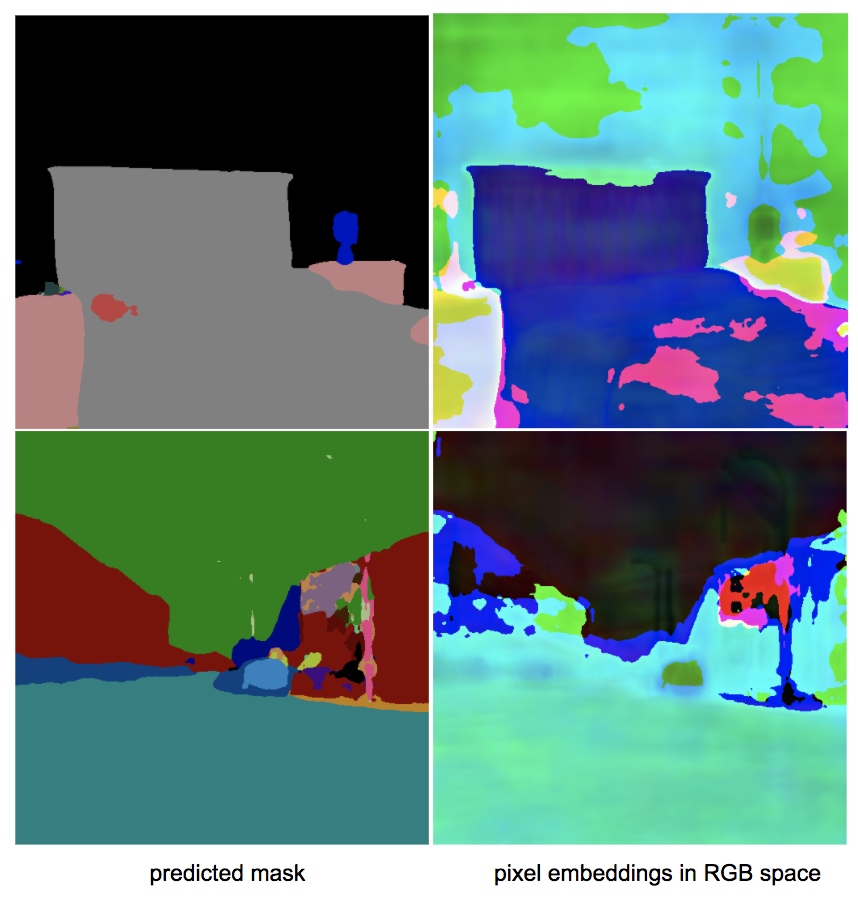}
    \vspace{-2mm}
    \caption[]{\textbf{Pixel Embeddings in RGB Space} Examples of predicted segmentation mask for ADE20k dataset (left). Pixel embeddings are projected into 3D space and transformed to RGB space (right).}
    \label{fig:RGB_embed}
    \vspace{-2mm}
\end{figure}

\begin{figure*}[t]
    \centering
    \includegraphics[height=1.0\linewidth]{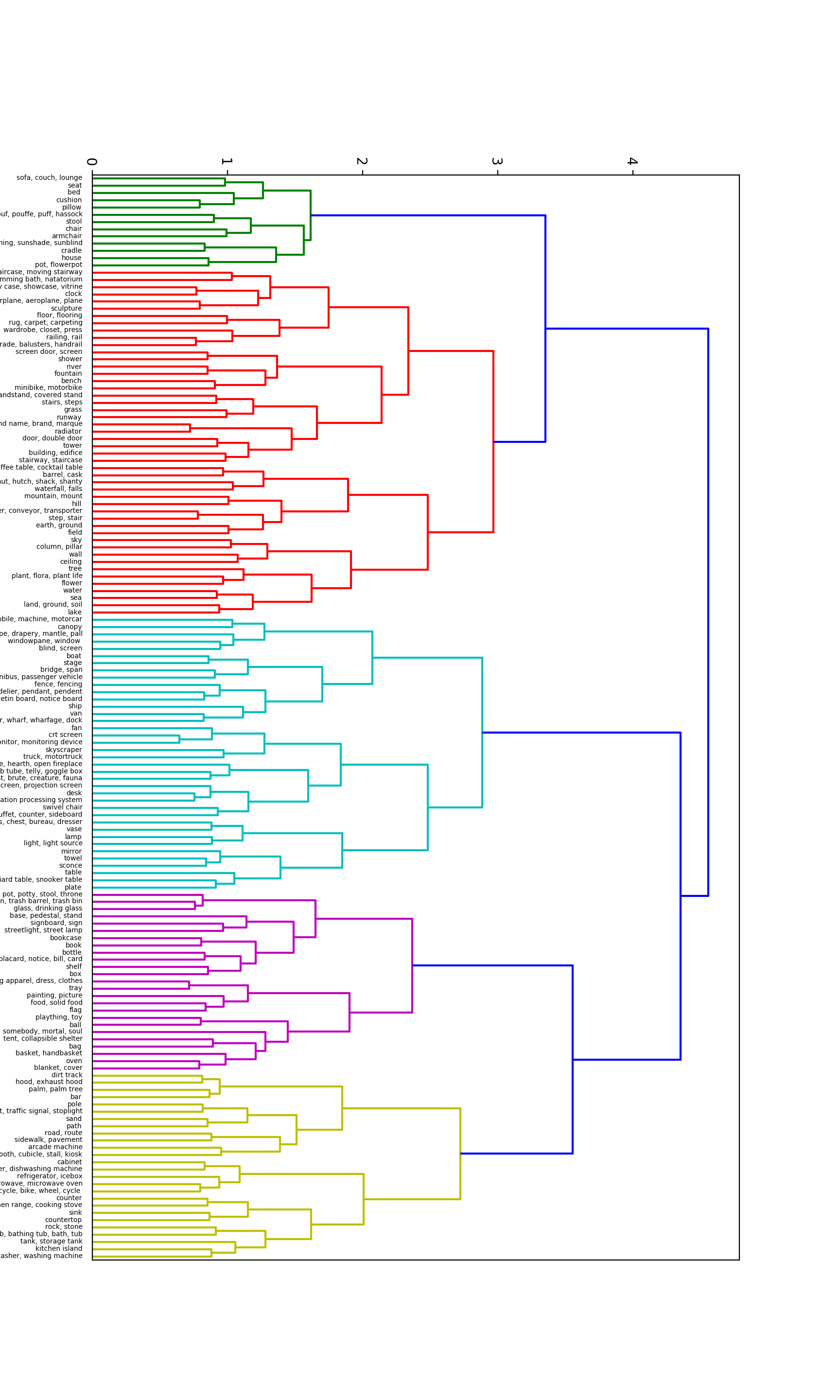}
    \vspace{-2mm}
    \caption[]{\textbf{Similarities in Class Embeddings}: Agglomerative clustering for ADE20k classes based on class embeddings learned by our ESS approach. We observe some of the semantically similar classes clustered together. For example, green sub-tree has couch, seat, bed, pillow, hassock, cushion, chair, stool and cradle classes clustered together. These classes often occur together in a bedroom or drawing room scene and are used for sitting or sleeping. In the yellow sub-tree towards bottom, we notice kitchen appliances clustered together.}
    \label{fig:ade20k_clusters}
    \vspace{-2mm}
\end{figure*}

\begin{figure*}[t]
    \centering
    \includegraphics[height=1.0\linewidth]{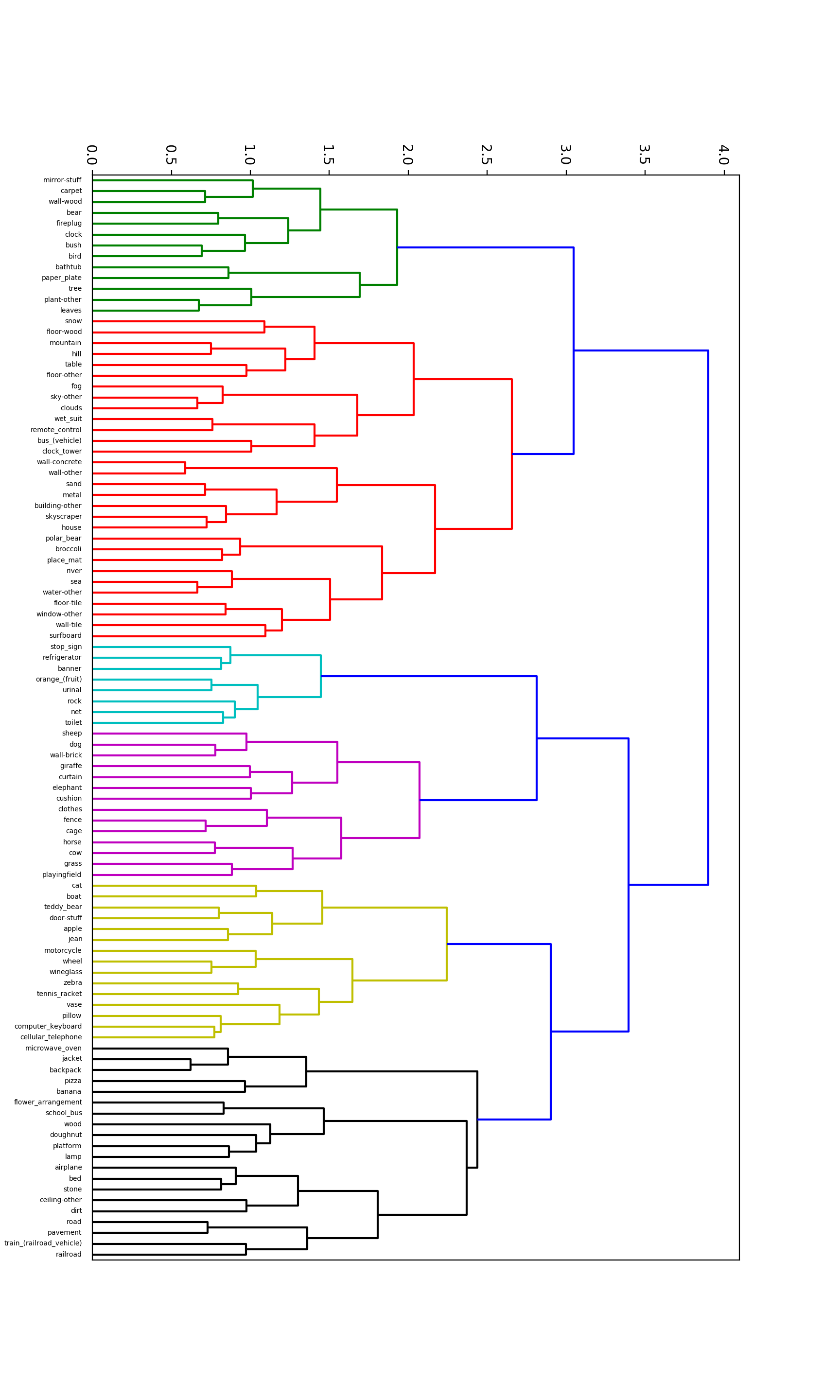}
    \vspace{-2mm}
    \caption[]{\textbf{Similarities in Class Embeddings}: Agglomerative clustering for COCO+LVIS classes based on class embeddings learned by our ESS approach}
    \label{fig:lvis}
    \vspace{-2mm}
\end{figure*}

\begin{figure*}[htb!]
    \centering
    \includegraphics[height=1.0\linewidth]{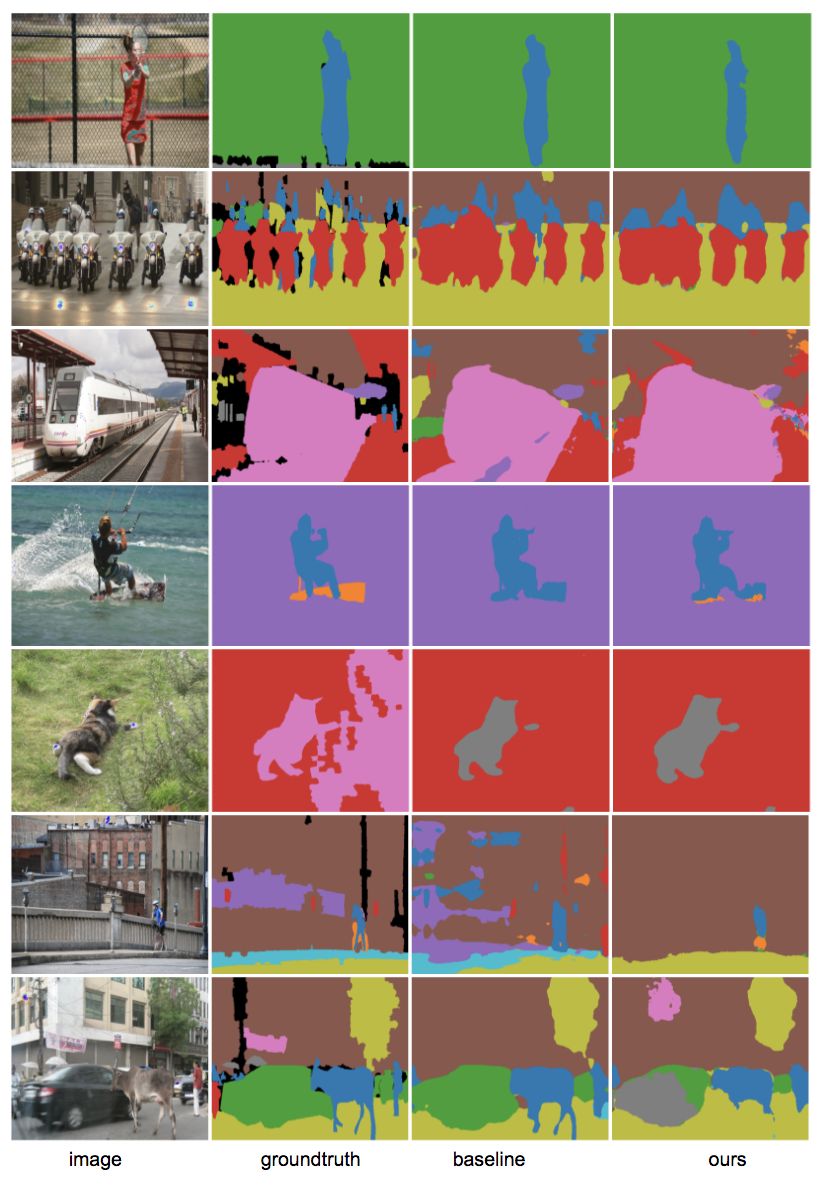}
    \vspace{-2mm}
    \caption[]{Qualitative results for COCO-Stuff10k dataset.}
    \label{fig:QR2}
    \vspace{-2mm}
\end{figure*}
\begin{figure*}[htb!]
    \centering
    \includegraphics[height=1.0\linewidth]{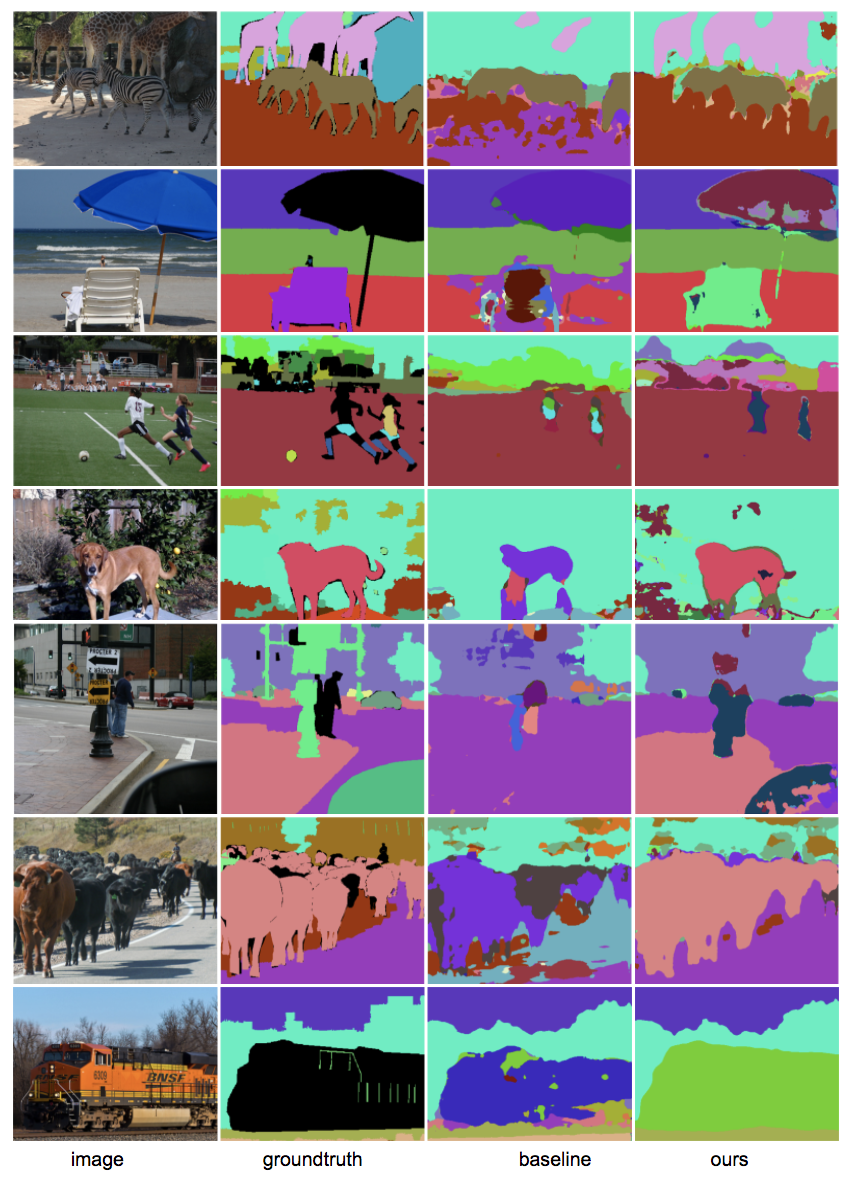}
    \vspace{-2mm}
    \caption[]{Qualitative results for COCO+LVIS dataset. Black color denotes the unlabelled pixels.}
    \label{fig:QR1}
    \vspace{-2mm}
\end{figure*}

\section{Detailed Qualitative Results}

We report more qualitative results for two most challenging datasets $\ie$ COCO-Stuff10k and COCO+LVIS in Figure \ref{fig:QR2} and \ref{fig:QR1} respectively.

\end{alphasection}
\end{document}